%% file: main.tex
\documentclass[runningheads]{llncs}
\usepackage[T1]{fontenc}
\usepackage{graphicx}
\usepackage{multirow}

\usepackage{orcidlink}
\usepackage{hyperref}

\begin{document}
\title{Domain-Specialized Interactive Segmentation Framework for Meningioma Radiotherapy Planning}
%
\titlerunning{Domain-Specialized Interactive Segmentation Framework for Meningioma RT}
%

\author{\orcidlink{0000-0001-7489-5829}Junhyeok Lee\inst{1}\textsuperscript{\dag} \and
\orcidlink{0009-0002-4719-3180}Han Jang\inst{2}\textsuperscript{\dag} \and
\orcidlink{0000-0002-5175-3307}Kyu Sung Choi\inst{2}\textsuperscript{\ddag}}

\authorrunning{Lee et al.}

\institute{Seoul National University of Medicine, Seoul, Republic of Korea \and
Seoul National University Hospital, Seoul, Republic of Korea}

\maketitle              

\renewcommand{\thefootnote}{\dag}
\footnotetext{These authors contributed equally to this work.}
\renewcommand{\thefootnote}{\ddag}
\footnotetext{Corresponding author. Email: ent1127@snu.ac.kr}
\renewcommand{\thefootnote}{\arabic{footnote}}

\begin{abstract}

Precise delineation of meningiomas is crucial for effective radiotherapy (RT) planning, directly influencing treatment efficacy and preservation of adjacent healthy tissues. While automated deep learning approaches have demonstrated considerable potential, achieving consistently accurate clinical segmentation remains challenging due to tumor heterogeneity. Interactive Medical Image Segmentation (IMIS) addresses this challenge by integrating advanced AI techniques with clinical input. However, generic segmentation tools, despite widespread applicability, often lack the specificity required for clinically critical and disease-specific tasks like meningioma RT planning. To overcome these limitations, we introduce Interactive-MEN-RT, a dedicated IMIS tool specifically developed for clinician-assisted 3D meningioma segmentation in RT workflows. The system incorporates multiple clinically relevant interaction methods, including point annotations, bounding boxes, lasso tools, and scribbles, enhancing usability and clinical precision. In our evaluation involving 500 contrast-enhanced T1-weighted MRI scans from the BraTS 2025 Meningioma RT Segmentation Challenge, Interactive-MEN-RT demonstrated substantial improvement compared to other segmentation methods, achieving Dice similarity coefficients of up to 77.6\% and Intersection over Union scores of 64.8\%. These results emphasize the need for clinically tailored segmentation solutions in critical applications such as meningioma RT planning. The code is publicly available at: \url{https://github.com/snuh-rad-aicon/Interactive-MEN-RT}


\end{abstract}

\input{introduction}
\input{methods}
\input{results}

\input{conclusion}

\begin{credits}

\subsubsection{\discintname}
The authors declare that they have no competing interests.
\subsubsection{Note}
This is the author’s accepted manuscript. The final authenticated version is available in 
\textit{Lecture Notes in Computer Science, vol. 16126, pp. XX–YY. Springer, Cham (2026).}
\url{https://doi.org/10.1007/978-3-032-05479-1_4}
\end{credits}


\bibliographystyle{splncs04_unsrt}
\bibliography{refs}
\end{document}

%% file: introduction.tex
\section{Introduction}

\begin{figure*}
    \centering
    \includegraphics[width=\textwidth]{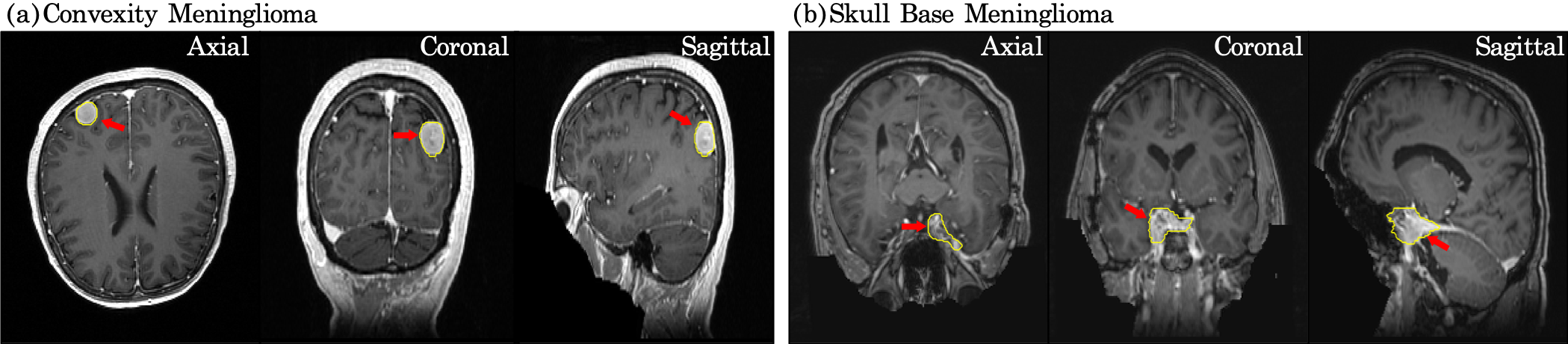}
    \caption{Representative meningioma cases illustrating anatomical diversity. (a) Convexity meningioma with clear margins; (b) Skull base meningioma encasing structures.}
    \label{fig:example}
\end{figure*}

Meningiomas, the most common primary intracranial tumors, frequently require radiotherapy~(RT) as a primary or adjuvant treatment modality. The success of RT is critically contingent upon the precise delivery of radiation to the target volume while minimizing exposure to adjacent healthy brain structures and organs at risk~(OARs)~\cite{goldsmith2018estro}. Accurate delineation of the gross tumor volume~(GTV) is therefore a cornerstone of effective RT planning. However, manual segmentation of meningiomas from medical imaging data—typically multi-parametric Magnetic Resonance Imaging~(MRI)—is a labor-intensive and technically demanding task. It is susceptible to substantial inter- and intra-observer variability, even among experienced radiation oncologists and neuroradiologists \cite{mabray2015variability}. This inconsistency is primarily attributable to the anatomical diversity and intrinsic heterogeneity of meningiomas, including variations in tumor location, shape, and imaging features such as cystic degeneration, necrosis, or calcification.

The anatomical complexity and heterogeneity of meningiomas, particularly those at the skull base or near intricate neurovascular structures, pose significant challenges to accurate boundary delineation. Fig.~\ref{fig:example} illustrates representative examples of meningiomas in distinct intracranial locations, highlighting the marked variability in tumor morphology, size, and relationship to adjacent anatomical structures. As depicted, convexity meningiomas typically exhibit well-defined margins and limited involvement with critical neurovascular anatomy. In contrast, skull base or ventricular meningiomas often abut or encase major vessels and cranial nerves, rendering manual segmentation challenging even for experts. These examples underscore the need for advanced segmentation tools that can robustly accommodate both the anatomical diversity and internal complexity of meningiomas encountered in radiotherapy planning.

To address these limitations, automated segmentation approaches, particularly those leveraging deep learning, have garnered considerable attention. Convolutional Neural Networks~(CNNs), especially U-Net-based architectures, have demonstrated state-of-the-art performance in diverse medical image segmentation tasks \cite{isensee2021nnunet,labella2024brats}. Nevertheless, the reliability of fully automated models remains suboptimal for cases involving heterogeneous tumor appearances or complex anatomical locations. Consequently, a persistent need for clinician oversight remains, especially in safety-critical applications like RT planning.

Interactive Medical Image Segmentation~(IMIS) frameworks have emerged as a promising solution, integrating artificial intelligence~(AI) with clinician expertise to facilitate more accurate and efficient tumor delineation \cite{cheng2024interactive}. The advent of foundation models such as the Segment Anything Model~(SAM) \cite{kirillov2023segmentanything,ravi2024sam} initially propelled progress in this domain, but early iterations were constrained by their two-dimensional scope. Recent advancements—including SAM-Med3D~\cite{wang2023sammed3d}, MedSAM2~\cite{ma2025medsam2}, and the nnInteractive framework \cite{isensee2025nninteractive}—have enabled interactive segmentation in volumetric medical images, fostering more effective clinician–AI collaboration. Despite these advances, many general-purpose IMIS tools treat the initial automated segmentation merely as a preliminary estimate, which may be insufficient for RT planning, where even minor errors can have significant dosimetric consequences.

In this study, we present Interactive-MEN-RT, a domain-specific and highly accurate IMIS tool tailored for meningioma delineation in the RT workflow. Built upon the robust nnU-Net V2 architecture~\cite{isensee2021nnunet,isensee2021nnunetv2} and enhanced with a bespoke interactive training module~\cite{isensee2025nninteractive}, our system is engineered to iteratively refine tumor segmentations in direct response to clinician input, thereby aligning the final contours with expert judgment. Our contributions are twofold:
(1) We show that adapting IMIS models to the meningioma RT domain, using task-specific fine-tuning and clinician-guided iterative refinement, leads to significant improvements over general-purpose baselines like SAM-Med3D and nnInteractive in both segmentation accuracy and clinical usability.
(2) We provide comprehensive validation demonstrating that, in the context of meningioma RT, specialized, domain-adapted models are essential to achieve the requisite accuracy and reliability for clinical use, with consistent gains observed across all forms of user interaction.

%% file: methods.tex
\section{Methods}

\begin{figure*}[!ht]
    \centering    \includegraphics[width=\textwidth]{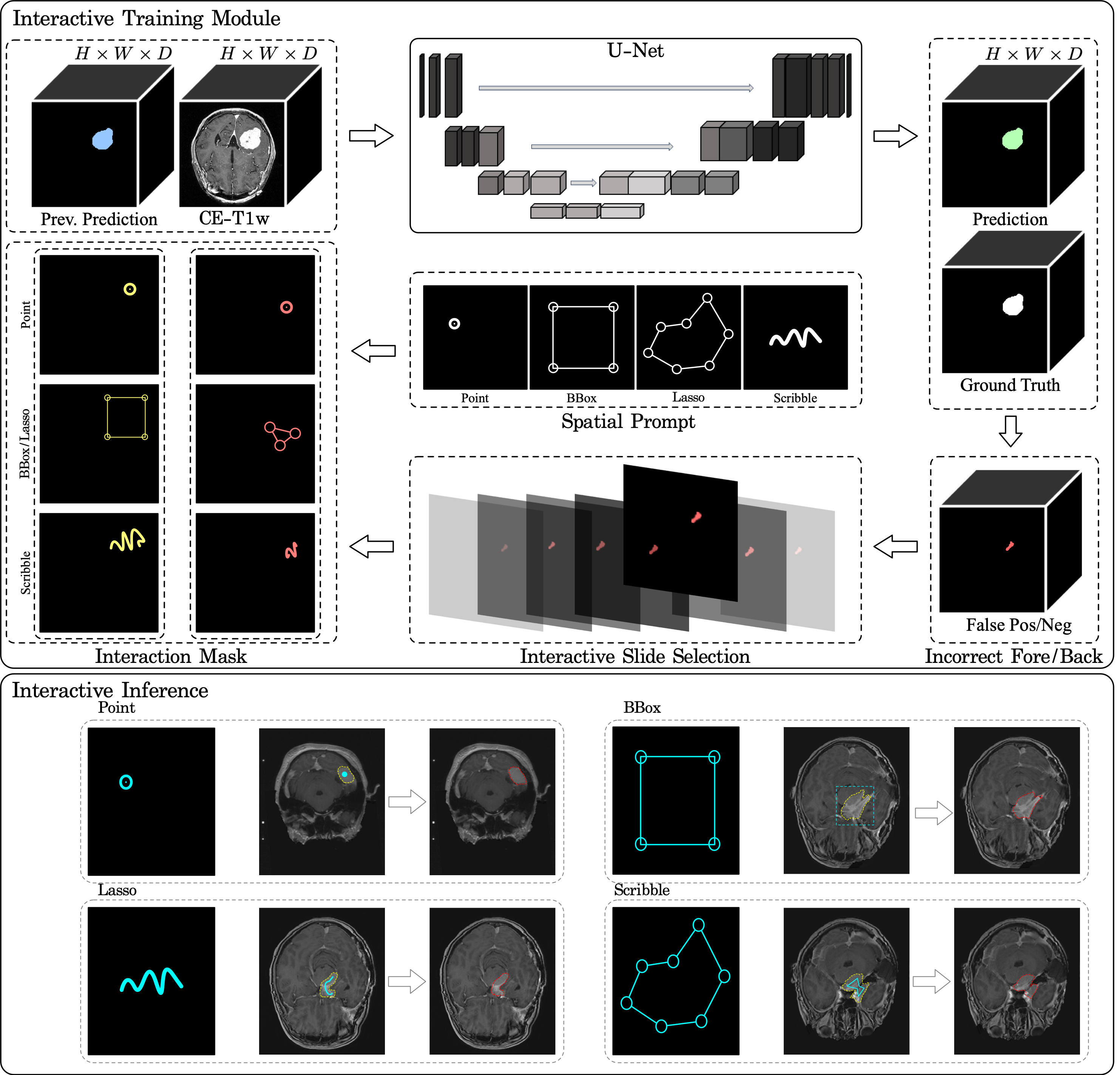}
    \caption{An overview of the Interactive-MEN-RT for meningioma gross tumor volume segmentation with interactive prompts.}
    \label{fig:overview}
\end{figure*}

This section presents the methodology of the Interactive-MEN-RT system for accurate meningioma segmentation in radiotherapy planning. It describes the data preprocessing pipeline, the U-Net–based model architecture, the integration and simulation of interactive user prompts, and an ablation study evaluating the impact of transfer learning~(TL). An overview of the entire pipeline is illustrated in Fig.~\ref{fig:overview}.

\subsection{Data Preprocessing and Preparation}

For our experimental validation, we exclusively utilized the training set of the BraTS 2025 Meningioma RT Segmentation Challenge dataset \cite{labella2024brats}. This comprehensive dataset comprises 500 samples specifically curated for meningioma Gross Tumor Volume~(GTV) segmentation in brain MRI. The dataset exclusively uses 3D contrast-enhanced T1-weighted~(CE-T1w) images and preserves extracranial structures, with patient-identifying facial features removed via defacing techniques to ensure anatomical integrity. The entire dataset was partitioned into 400 samples for training and 100 for validation to facilitate robust model evaluation.

Our preprocessing pipeline adapted robust protocols from nnU-Net: (1) intensity normalization via Z-score standardization, (2) resampling to 1 mm isotropic voxel spacing, (3) spatial cropping to eliminate excessive background while preserving anatomical context, and (4) data augmentation—including rotation, scaling, elastic deformation, and intensity variations—to enhance model robustness.

\subsection{Network Architecture}

The architecture of Interactive-MEN-RT is based on the principles of nnInteractive \cite{isensee2025nninteractive}, a 3D promptable segmentation framework for volumetric medical image analysis. Consistent with nnInteractive's design, we employ an U-Net-based architecture over Transformer alternatives, building upon the nnU-Net framework \cite{vaswani2017attention,isensee2021nnunet,isensee2021nnunetv2} and using the Residual Encoder~(ResEnc-L) configuration as its backbone. The architecture incorporates nnInteractive's sophisticated prompt-processing mechanisms, adapted for the specific requirements of meningioma segmentation to efficiently integrate user guidance. We leverage nnInteractive's multi-channel input design, which comprises three key components: (1) the original CE-T1w MRI data, (2) previous segmentation results, and (3) interactive guidance signals. Following nnInteractive's methodology, the network supports a comprehensive range of spatial prompt types, including points, bounding boxes, lasso selections, and scribbles \cite{isensee2025nninteractive,wong2024scribbleprompt}. Each prompt type is encoded in two separate input channels (positive and negative) within the interactive guidance signals.

We optimized the combination of loss functions for the binary task of differentiating tumor from background in a radiotherapy planning context. The final layer is configured for binary segmentation (tumor vs. background) with a sigmoid activation function, optimized using a combination of Dice loss and Cross-Entropy loss~(DiceCELoss) to balance overlap accuracy with voxel-wise precision.

\subsection{Interaction Prompts}

\subsubsection{User Interaction Types} 
Drawing from advancements in IMIS systems such as nnInteractive \cite{isensee2025nninteractive}, our module supports intuitive 2D prompts on a standard axial view to guide 3D segmentation. The supported prompt types include points for correcting localized errors with positive foreground and negative background clicks; bounding boxes to define broad regions of interest; lasso selections as closed-loop contours for precise delineation of irregular tumor boundaries; and scribbles as free-form lines to indicate larger regions for inclusion or exclusion. These user interactions are converted into spatial maps that indicate areas of interest or correction, which are then combined with the original MRI to incorporate clinician expertise directly into the segmentation refinement process.

\subsubsection{Training-time Simulation of Prompts} 
During training, we simulated realistic user interactions to improve model robustness and clinical relevance. For each interaction type, including point, bounding box, scribble, and lasso prompts, we employed dedicated sampling strategies that emulate clinical correction workflows while introducing stochastic variations. For point prompts, we randomly sampled 1–2 positive points within the tumor, randomizing their location, size, and number. Bounding box prompts involved generating a single box on a tumor-containing slice with a random margin, often including jitter or size variation. Scribble prompts were created by connecting 2–8 random points within the tumor on a slice, with randomized order and added jitter or wavy effects to mimic freehand drawing. Finally, lasso prompts simulated polygons by sampling 4–12 jittered points along the tumor boundary on a selected slice, ensuring valid closed contours. Across all prompt types, slice selection was weighted by tumor area, and all randomizations captured the diversity of real-world clinical interactions.

\subsection{Prompt Encoding}
All user prompts~(points, boxes, lassos, scribbles) are encoded as two additional channels: one for positive (foreground) and one for negative (background) interactions. Each channel matches the spatial dimensions of the input image and is normalized to the [0, 1] range. These prompt channels are concatenated with the original image to form the network input, allowing the model to flexibly incorporate user guidance.

\subsection{Ablation Study: Transfer Learning}
To evaluate the effect of transfer learning, we compared two model initialization strategies: (1) training the network from scratch with random initialization, and (2) using pre-trained nnInteractive weights for transfer learning. Both models were trained and evaluated under the same protocol and dataset settings.

\subsection{Implementation Details}
All models were implemented in PyTorch 1.13 and trained on a NVIDIA A6000 GPU with 48 GB of memory. Training was performed using 3D patches of size 128×128×128 and a batch size of 8. The optimizer was SGD with Nesterov momentum, an initial learning rate of 1e-2, and a polynomial decay schedule. Data augmentation included random rotations~(±15°), scaling~(0.9–1.1×), elastic deformation, and intensity shifts~(±10\%). Consistent preprocessing and evaluation protocols were used for all experiments.

\begin{table}[t]
\caption{Interactive segmentation performance by prompt type (mean $\pm$ SD).}
\label{tab:int_prompt}
\centering
\setlength{\tabcolsep}{6pt}
\renewcommand{\arraystretch}{1.1}
\small
\begin{tabular}{l@{\hskip 6pt}lcc}
\hline
\textbf{Prompt} & \textbf{Method} & \textbf{DSC (\%)} & \textbf{IoU (\%)}\\
\hline
\multirow{1}{*}{None}
 & nnUNet                                          & $65.5\pm25.1$ & $53.1\pm23.9$\\
\hline
\multirow{5}{*}{Point}
 & MedSAM2\,\cite{ma2025medsam2}                  & $10.1\;\pm\;9.2$ & $5.1\;\pm\;5.0$\\
 & SAM-Med3D\,\cite{wang2023sammed3d}             & $74.9\pm21.9$ & $60.4\pm20.5$\\
 & nnInteractive                                  & $69.7\pm22.2$ & $57.2\pm22.3$\\
 & Interactive-MEN-RT (scratch)                   & $\mathbf{75.5\pm16.1}$ & $\mathbf{62.8\pm17.2}$ \\
 & Interactive-MEN-RT (TL)                        & $72.8\pm20.4$ & $60.5\pm20.6$\\
\hline
\multirow{4}{*}{BBox}
 & MedSAM2\,\cite{ma2025medsam2}                  & $63.2\pm21.5$ & $49.1\pm19.6$\\
 & nnInteractive                                  & $50.1\pm16.4$ & $35.0\pm14.9$\\
 & Interactive-MEN-RT (scratch)                   & $76.0\pm13.0$ & $62.9\pm15.3$\\
 & Interactive-MEN-RT (TL)                        & $\mathbf{77.6\pm11.2}$ & $\mathbf{64.6\pm13.7}$\\
\hline
\multirow{3}{*}{Lasso}
 & nnInteractive                                  & $61.8\pm21.2$ & $47.9\pm20.9$\\
 & Interactive-MEN-RT (scratch)                   & $\mathbf{77.5\pm13.0}$ & $\mathbf{64.8\pm15.1}$\\
 & Interactive-MEN-RT (TL)                        & $63.6\pm30.2$ & $52.6\pm27.5$\\
\hline
\multirow{3}{*}{Scribble}
 & nnInteractive                                  & $73.9\pm17.7$ & $61.2\pm18.7$\\
 & Interactive-MEN-RT (scratch)                   & $\mathbf{76.2\pm15.0}$ & $\mathbf{63.5\pm16.5}$\\
 & Interactive-MEN-RT (TL)                        & $69.7\pm25.0$ & $58.0\pm23.7$\\
\hline
\end{tabular}
\label{tab:seg_performance}
\end{table}

\begin{figure*}[!ht]
  \centering
  \includegraphics[width=\textwidth]{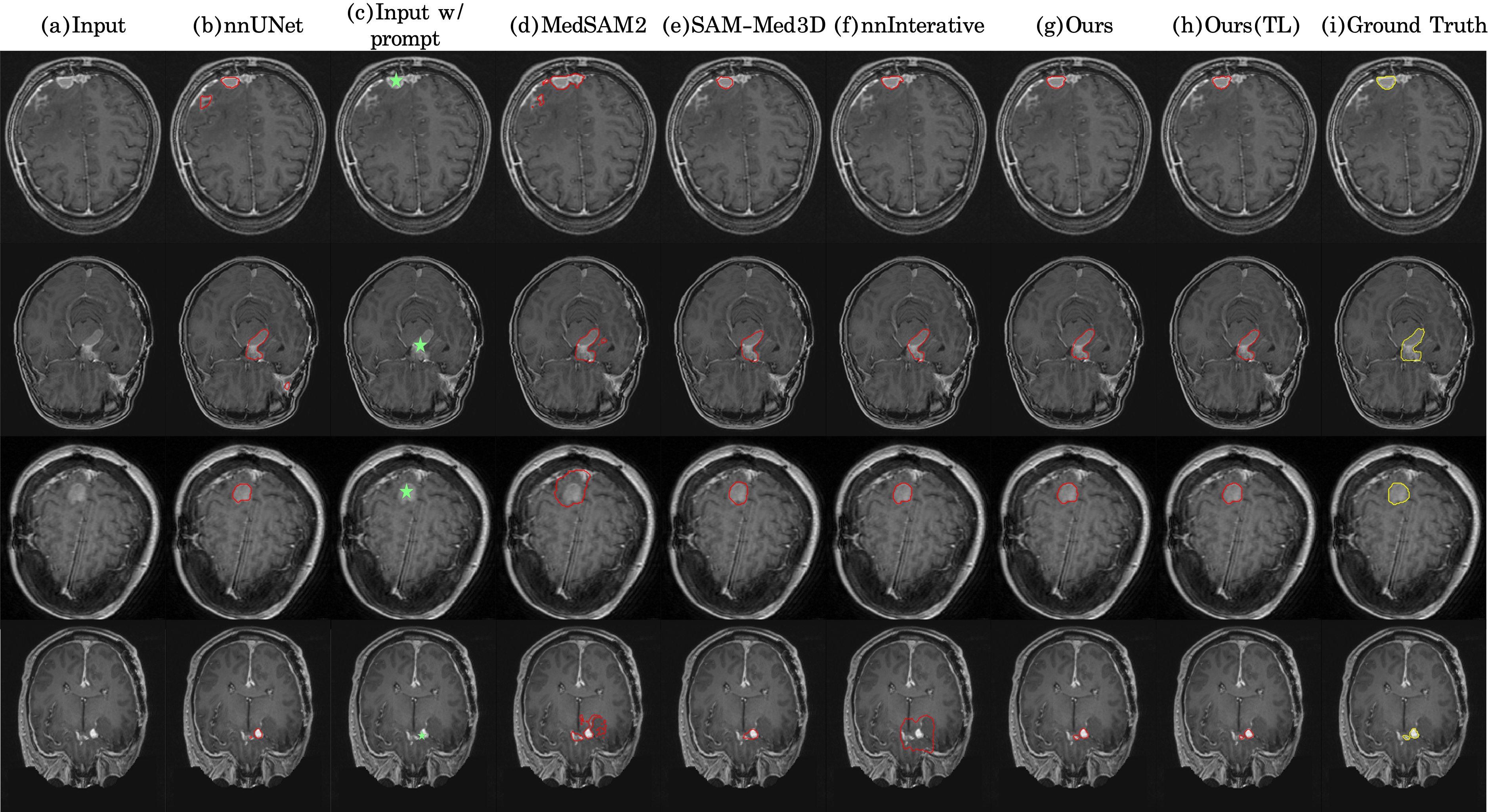}
  \caption{Qualitative segmentation overlays for each method and the ground truth under point prompt interaction settings.}
  \label{fig:seg_overlay}
\end{figure*}

\subsection{Evaluation Metrics and Baselines}

Segmentation accuracy was evaluated on the validation set using two standard metrics: the Dice Similarity Coefficient~(DSC) and Intersection over Union~(IoU). We compared the performance of Interactive-MEN-RT with several state-of-the-art interactive segmentation frameworks, including nnInteractive \cite{isensee2025nninteractive}, SAM-Med3D \cite{wang2023sammed3d}, and MedSAM2 \cite{ma2025medsam2}, as well as an ablation variant that distinguished between training from scratch and transfer learning. To compare with an automatic segmentation model, we additionally included nnUNet \cite{isensee2021nnu} as a strong baseline due to its proven generalization capability across diverse medical segmentation tasks.
To evaluate the interactive methods, user interactions were simulated by sampling prompts from ground truth tumor regions on a per-case basis.

%% file: results.tex
\section{Results}

\subsection{Qualitative Assessment by Prompt Type}
As shown in the interactive inference results in Fig.~\ref{fig:overview}, Interactive-MEN-RT consistently produces accurate and robust segmentations across diverse prompt types. 
The visual examples demonstrate that Interactive-MEN-RT consistently achieves high-quality segmentations regardless of the prompt strategy employed. Interactive-MEN-RT adeptly handles various input modalities, from point prompts for initial delineations to bounding box prompts for refining complex regions. This adaptability is crucial for real-world clinical scenarios where diverse tumor characteristics and user preferences necessitate flexible interaction.

\begin{figure*}[!ht]
  \centering
  \includegraphics[width=\textwidth]{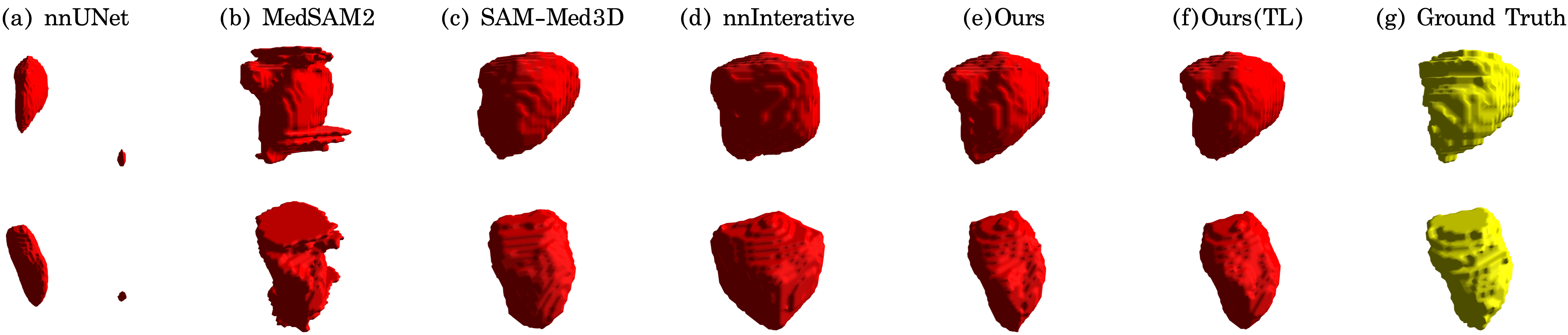}
  \caption{Qualitative 3D segmentations for each method and the ground truth.}
  \label{fig:seg_overlay_3D}
\end{figure*}

\subsection{Interactive Segmentation Performance}
Table~\ref{tab:seg_performance} presents a comparative performance analysis of Interactive-MEN-RT against leading interactive segmentation models, including SAM-Med3D \cite{wang2023sammed3d}, MedSAM2 \cite{ma2025medsam2}, and nnInteractive \cite{isensee2025nninteractive}. The analysis evaluates segmentation accuracy~(DSC and IoU) across various prompt types: points, bounding boxes, lassos, and scribbles. This comparison also serves as an ablation study, highlighting the impact of transfer learning within the Interactive-MEN-RT framework.


Interactive-MEN-RT consistently demonstrated performance superior or comparable to the baseline methods across all prompt types. With point-based prompts, the scratch-trained Interactive-MEN-RT model yielded the highest DSC~(75.5\% $\pm$ 16.1), outperforming all baselines, while its transfer learning~(TL) variant achieved a leading IoU score~(60.5\% $\pm$ 20.6). For bounding box interactions, the TL model achieved the highest DSC~(77.6\% $\pm$ 11.2) and IoU~(64.6\% $\pm$ 13.7), with the scratch-trained model also showing strong performance. In the case of lasso prompts, the scratch-trained model again led in both DSC~(77.5\% $\pm$ 13.0) and IoU~(64.8\% $\pm$ 15.1), substantially outperforming both nnInteractive and the TL variant. Similarly, for scribble-based corrections, the scratch model obtained the highest DSC~(76.2\% $\pm$ 15.0) and IoU~(63.5\% $\pm$ 16.5).

\subsection{Comparative Qualitative Analysis of Segmentation Methods}
Fig.~\ref{fig:seg_overlay} presents a qualitative comparison of segmentation results obtained under a single point prompt on representative meningioma cases. The figure facilitates a direct visual comparison of outputs from MedSAM2, SAM-Med3D, nnInteractive, Interactive-MEN-RT~(trained from scratch), and Interactive-MEN-RT~(with transfer learning) against the ground truth annotations. Each row corresponds to a distinct meningioma case, strategically chosen to exemplify the diversity of tumor locations relevant to RT planning: convexity, skull base, falx/parasagittal, and ventricular regions. This presentation of qualitative results across varied clinical scenarios effectively demonstrates the robustness and adaptability of Interactive-MEN-RT.

As shown in Fig.~\ref{fig:seg_overlay_3D}, volumetric renderings of predicted segmentations illustrate the spatial coherence and completeness of Interactive-MEN-RT compared to baseline models, offering a more comprehensive perspective on tumor coverage that may be particularly helpful for irregular or deeply situated lesions in anatomically complex regions such as the skull base and ventricles.

%% file: conclusion.tex
\section{Discussion}
This study demonstrates the clinical and technical value of Interactive-MEN-RT, a specialized interactive segmentation system designed for meningioma radiotherapy planning.
It enables clinicians to delineate target volumes with minimal input, reducing inter-observer variability and streamlining RT workflows.
Its robustness across diverse prompt types—including points, bounding boxes, lassos, and scribbles—ensures flexibility in accommodating varying user preferences and clinical scenarios.
Importantly, the ability to achieve high accuracy with minimal input makes it well-suited for time-sensitive clinical workflows.

Our comparative analysis demonstrates that Interactive-MEN-RT consistently achieves superior or competitive performance compared to established baseline models, such as SAM-Med3D~\cite{wang2023sammed3d}, MedSAM2~\cite{ma2025medsam2}, and nnInteractive~\cite{isensee2025nninteractive}. The most substantial performance improvements were observed with point and lasso prompts, emphasizing the effectiveness of our prompt encoding and training methodologies.
These results highlight the value of domain-specific models in safety-critical tasks where general-purpose tools may fall short.

Models trained from scratch on task-specific data often surpassed transfer learning, especially with fine-grained prompts.
Although nnInteractive generalizes well, it underperforms in capturing lesion-specific details compared to scratch-trained models.
This limitation likely arises from the model's optimization towards particular prompt types or anatomical contexts, restricting its adaptability during fine-tuning.
Transfer learning remains useful when data or compute is limited, especially for simple prompts like boxes.
A hybrid strategy combining transfer learning and domain-specific training is recommended for precision-critical applications.

This study has several limitations. First, user interactions were simulated rather than obtained directly from clinicians, potentially limiting the representation of the diverse and complex nature of real-world clinical usage. Second, the evaluation was limited to a single dataset~\cite{labella2024brats}, 
raising concerns about generalizability.
Finally, prospective clinical studies and comprehensive user experience assessments are needed to confirm the system’s practical utility and usability in clinical settings.

\section{Conclusion}
In conclusion, Interactive-MEN-RT presents a promising interactive segmentation solution tailored for meningioma radiotherapy planning, achieving strong performance across various prompt types. Its ability to deliver accurate segmentation with minimal input, combined with favorable comparisons against established baselines, suggests strong potential to enhance efficiency, consistency, and safety in clinical radiotherapy workflows. These findings underscore the value of utilizing specialized, disease-specific models in safety-critical medical settings, as they may better address the limitations of general-purpose frameworks and align more closely with clinical practice demands.